\documentclass[11pt]{article}
\usepackage{authblk}
\usepackage{acl}
\usepackage{float}
\usepackage{graphicx}

\usepackage{times}
\usepackage{latexsym}

\usepackage[T1]{fontenc}

\usepackage[utf8]{inputenc}

\usepackage{microtype}

\usepackage{cleveref}
\usepackage{booktabs}
\usepackage[frozencache=true,cachedir=minted-cache]{minted} 
\usepackage{graphicx}
\usepackage[inkscapearea=drawing]{svg}
\usepackage[colorinlistoftodos]{todonotes}
\usepackage{multirow}
\usepackage{pbox}
\usepackage{xspace}
\usepackage{stfloats}

\usepackage{url}
\usepackage{tikz}
\usepackage[breaklinks]{hyperref}

\newcommand{\square}{{\textsc{UKP-SQuARE}}\xspace}

\def\checkmark{\tikz\fill[scale=0.3](0,.35) -- (.25,0) -- (1,.7) -- (.25,.15) -- cycle;}

\title{$\vcenter{\hbox{\includegraphics[width=0.35in]{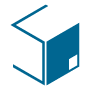}}}$~\square: An Online Platform for \\Question Answering Research}
\author{
\begin{minipage}[t]{\textwidth}
\centering
\normalsize
\bf
Tim Baumgärtner,$^{1}$
Kexin Wang,$^{1}$
Rachneet Sachdeva,$^{1}$
Max Eichler,$^{1}$
Gregor Geigle,$^{1}$
Clifton Poth,$^{1}$
Hannah Sterz,$^{1}$
Haritz Puerto,$^{1}$
Leonardo F. R. Ribeiro,$^{1}$
Jonas Pfeiffer,$^{1}$
Nils Reimers,$^{2}$
Gözde Gül Şahin,$^{3}$
Iryna Gurevych$^{1}$ \\
{\footnotesize \normalfont 
$^{1}$Ubiquitous Knowledge Processing Lab, Department of Computer Science, Technical University of Darmstadt \\
$^{2}$Hugging Face, $^{3}$Koç University
Computer Science and Engineering
Department \\
\url{www.ukp.tu-darmstadt.de}
} 
\end{minipage}
}
\begin{document}
\maketitle

\begin{abstract}
Recent advances in NLP and information retrieval have given rise to a diverse set of question answering tasks that are of different formats (e.g., extractive, abstractive), require different model architectures (e.g., generative, discriminative), and setups (e.g., with or without retrieval). Despite having a large number of powerful, specialized QA pipelines (which we refer to as Skills) that consider a single domain, model or setup, there exists no framework where users can easily explore and compare such pipelines and can extend them according to their needs. To address this issue, we present \square, an extensible online QA platform for researchers which allows users to query and analyze a large collection of modern Skills via a user-friendly web interface and integrated behavioural tests. In addition, QA researchers can develop, manage, and share their custom Skills using our microservices that support a wide range of models (Transformers, Adapters, ONNX), datastores and retrieval techniques (e.g., sparse and dense). \square is available on \url{https://square.ukp-lab.de}.\footnote{The code is available on \url{https://github.com/UKP-SQuARE/square-core}}
\end{abstract}

\section{Introduction}

Researchers in NLP have devoted significant resources to creating more powerful machine learning models for Question Answering (QA), and collecting high-quality QA datasets. Combined with the recent breakthroughs by large pretrained language models, we have witnessed rapid progress in the field across many different kinds of QA tasks~\cite{rogersqa21}. 

\begin{figure}[t]
    \centering
    
    \includegraphics[width=\columnwidth]{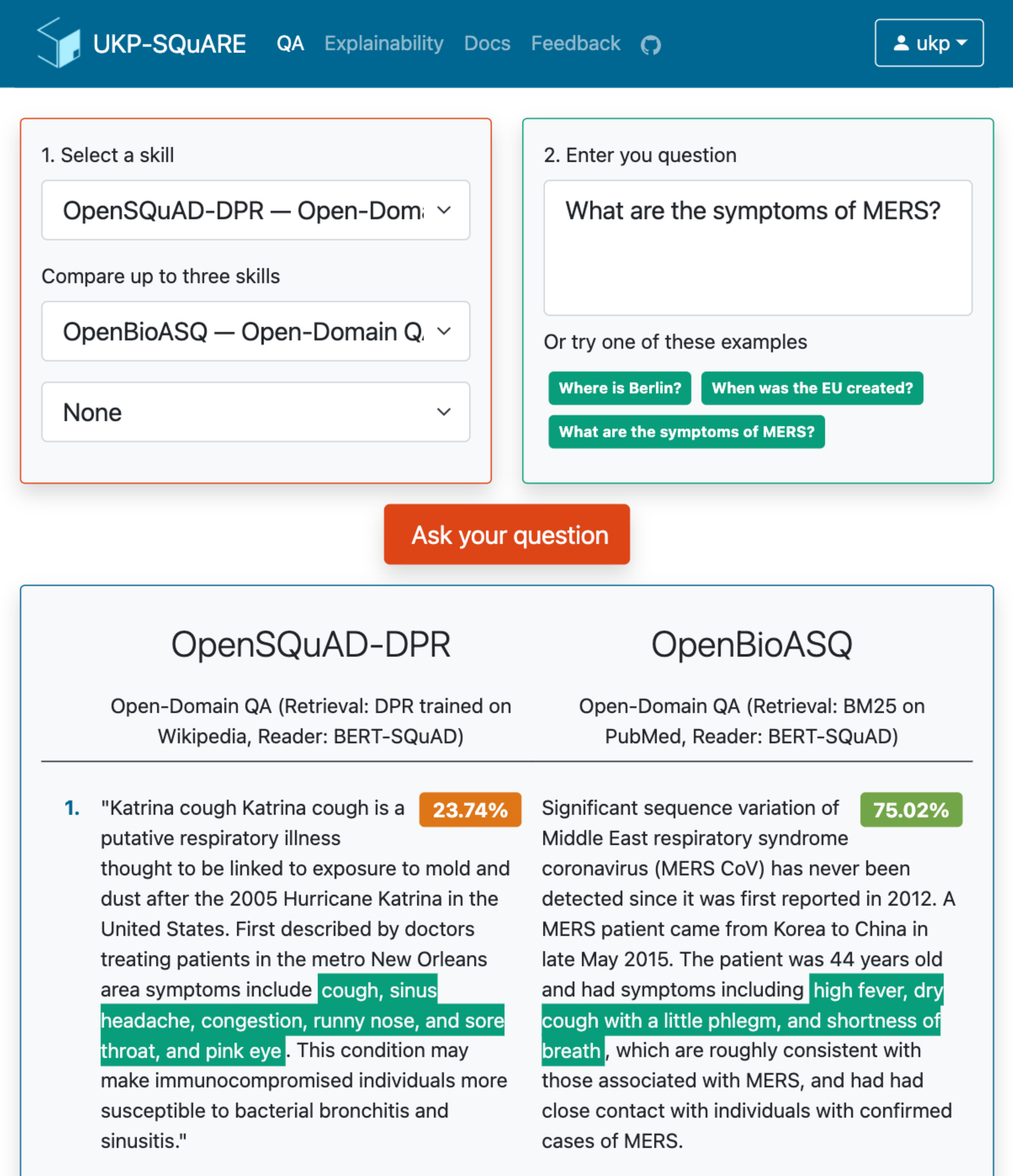}
    \caption{QA page of \square. The user selects a Skill (in this case, two open-domain Skills are selected), enters a question and then receives an answer.} 
    \label{fig:screenshot-question}
    \vspace{-4mm}
\end{figure}

The great variety in QA tasks has led to specialized, domain-specific models trained on a single QA format such as \textit{multiple choice}~\cite{lai2017race} (i.e., selecting the best answer out of multiple options), \textit{extractive}~\cite{rajpurkar2016squad} (i.e., finding the answer span in a context) and \textit{abstractive}~\cite{kocisky2018narrativeqa} (i.e., generating an answer that is not a contiguous span in the context). The format may influence the model architecture  (e.g., discriminative objective for \textit{multiple choice}, generative objective for \textit{abstractive}). Additionally, systems vary with how the context is provided. It can be given by the user, or retrieved from a Datastore which is commonly referred to as \textit{open-domain} or \textit{retriever-reader} setup~\cite{chen2017reading}. The retrieval mechanism can also be chosen from a set of sparse (e.g., BM25, ~\citeauthor{DBLP:conf/trec/RobertsonWJHG94}, \citeyear{DBLP:conf/trec/RobertsonWJHG94}) or dense (e.g., DPR,~\citeauthor{karpukhin-etal-2020-dense}, \citeyear{karpukhin-etal-2020-dense}) techniques.

The speed of progress in the field makes it essential for researchers to explore, compare, and combine these different QA components as quickly as possible to identify the strengths and weaknesses of the current state of the art. Even though there exists a number of powerful QA systems~\cite{dibia2020neuralqa,khashabi2020unifiedqa} and frameworks such as Haystack,\footnote{\url{https://deepset.ai/haystack}} those approaches focus only on one component (e.g., retrieval, QA format, domain), hence do not allow plug-and-play of different Datastores, domains, model architectures or retrieval techniques. This considerably limits their applicability and reusability across the diverse, rapidly progressing area of QA research, making it infeasible for researchers to quickly integrate novel models and QA pipelines. \par
To address this gap, we introduce \square, a flexible and extensible QA platform to enable users to easily implement, manage and share their custom QA pipelines, which we call \textbf{Skills}, using our microservices. As shown in Fig.~\ref{fig:screenshot-question}, \square also allows users to query and compare different \textbf{Skills} via an easy-to-use user interface and systematically analyze their strengths and weaknesses through integrated behavioural tests.\footnote{Screenshots for adding Skills, the outputs of different QA formats and behavioural tests are shown in Appendix~\ref{app:qa}.}

\section{\square}

\begin{figure*}[!htp]
    \centering
    \includegraphics[width=1\textwidth]{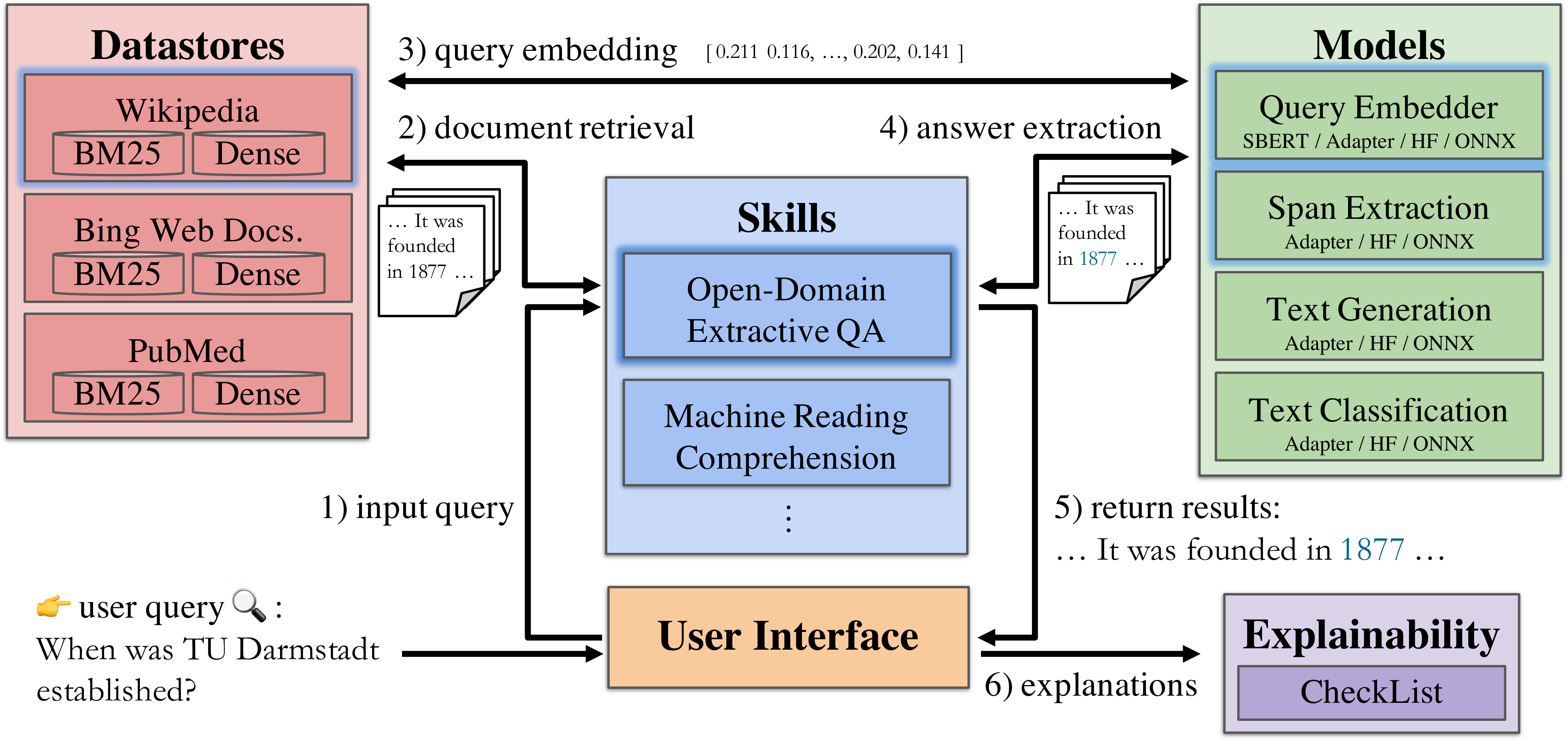}
    \caption{Overall architecture of \square, illustrating an open-domain, extractive QA Skill. (1) First a user selects a \textbf{Skill} and issues a query via the \textbf{User Interface}.
    (2) The selected QA Skill forwards the query to the respective \textbf{Datastore} for document retrieval. 
    (3) The \textbf{Datastore} gets the query embedding from the \textbf{Models}, uses it for semantic document retrieval and returns the top documents to the \textbf{Skill}. 
    (4) The \textbf{Skill} sends the query and retrieved documents to the reader model for answer extraction. 
    (5) Finally, the answers are shown to the user.
    (6) Optionally, the user can view the results of the predefined behavioural tests for the \textbf{Skill}. 
    } 
    \label{fig:architecture}
     \vspace{-4mm}
\end{figure*}

The system is implemented as a modern microservice architecture using Docker containers.\footnote{\url{https://docker.com}} The major components are \textbf{Skills}, \textbf{Datastores}, \textbf{Models}, \textbf{Explainability} and the \textbf{User Interface}.
The process flow across the components is illustrated in Fig.~\ref{fig:architecture} on an open-domain, extractive QA Skill. The central component of the system is the \textbf{Skill} that specifies how a user query is processed (e.g., which QA type, retrieval mechanism, model or adapter to be used in which order). The \textbf{Skill} leverages the other services for query execution.
\textbf{Datastores} hold multiple collections of documents with sparse indices, e.g., BM25~\cite{DBLP:conf/trec/RobertsonWJHG94} and dense indices, e.g., DPR~\cite{karpukhin-etal-2020-dense}, allowing fast and efficient retrieval of background knowledge in an extensible way. The \textbf{Model} service hosts numerous models, combined with Adapters \citep{pmlr-v97-houlsby19a, pfeiffer2020adapterhub}, to support a wide range of tasks such as text embedding (for queries in open-domain QA), sequence and token classification (for multiple-choice and extractive QA) and sequence-to-sequence generation (for abstractive QA).
The \textbf{Explainability} component performs behavioural tests on the deployed Skills for better understanding of the models. Details of each service are provided in the following sections. \par
Furthermore, while we host \square on our infrastructure and make it available for the community, we also provide the option to set up the system locally. Additionally, the \textbf{Datastores} and \textbf{Models} services are exposed via an API\footnote{{\small \url{https://square.ukp-lab.de/docs/}}}. 

\subsection{Skills} \label{sec:skill_api}

Skills define how the user query should be processed by the Datastores and Models components and how the respective answers are obtained. For question answering, this might involve retrieving background knowledge, extracting spans from context or selecting an answer from multiple choices. \par
Skills are not necessarily equivalent to a model trained on a dataset. Instead a Skill is more general and can use multiple models to arrive at an output. A Skill might work on a specialized domain (e.g. biomedical, movies, etc.) or a specific format (e.g. extractive, abstractive, etc.), but also combinations are possible. For example, a Skill could combine Wikipedia and a news based extractive reader model to answer factoid and news questions. The degree of specialization or generalization of a Skill is up to its developer. In \square the Skill only defines the pipeline, i.e., pre-processing, information retrieval or answer extraction/generation/classification. These steps are facilitated and executed by the usage of the other components: Models (\S \ref{sec:model_api}) and Datastores (\S \ref{sec:datastore_api}).

Importantly, Skills can be added to the system by the community. They can be added privately, thereby only giving a specific user access to it, or made public, allowing everyone to use it (\S \ref{sec:add_skill}). This allows great flexibility in the design of question answering pipelines, keeping implementation effort and required compute low, thereby democratizing the usage of  question answering models.\par

\subsection{Models}\label{sec:model_api}
The Models component is responsible for hosting NLP models required for document retrieval and answer extraction/generation tasks. Our platform supports a wide variety of models comprising HuggingFace (HF) Transformers~\cite{wolf-etal-2020-transformers}, Adapters, Sentence-Transformers~\cite{reimers-2019-sentence-bert}, and a limited selection of ONNX (Open Neural Network Exchange)~\cite{bai2019} models. Specifically, the inclusion of memory-efficient adapters in our platform allows having a variety of task-specific models while maintaining storage efficiency. Moreover, for faster inference, the high performance inference engine, ONNX Runtime\footnote{\url{https://github.com/microsoft/onnxruntime}} can be used for the ONNX models provided on our platform. 

The Models component comprises of two main services: \textbf{inference} and \textbf{management}. The inference service is responsible for loading models and getting predictions for the input queries. The management service allows the user to list, deploy, update and remove models (available on HF, Adapterhub and Sentence-Transformers) on the \square platform. This allows to deploy and query models beyond the ones we already provide, for example multilingual models. To maintain a scalable architecture, we host every deployed model in its separate Docker container and use Traefik\footnote{\url{https://traefik.io}} to route the user query to the specific model instance for inference. The inference service of the model API can be queried using the Skills (\S \ref{sec:skill_api}) as per the end-user's requirements.\par

\subsection{Datastores}\label{sec:datastore_api}

The Datastores are responsible for storing document collections as knowledge bases of QA Skills, supporting retrieval on these collections. Each Datastore contains a collection of documents and several indices of them for retrieval. The document collections are stored by an Elasticsearch\footnote{\url{https://elastic.co}} instance. Within one Datastore, the document collection is indexed by sparse or dense retrieval models. 

For sparse retrieval, we use BM25 provided by the Elasticsearch instance; for dense retrieval, we use dual-encoder neural networks~\cite{karpukhin-etal-2020-dense,xiong2021approximate} with Approximate Nearest Neighbor (ANN) indexing provided by Faiss~\cite{JDH17}. The Datastores are agnostic to the ANN methods. Among them, we use \textit{IndexIVFScalarQuantizer}~\cite{DBLP:journals/pami/JegouDS11} from Faiss as the default choice. For scalability, we maintain each dense-retrieval index within one Docker container and use Traefik to route the queries to the specific index. For each query using dense retrieval, the Datastores forward the query to the Models to get the query embedding (e.g., via the Query Embedder in Fig.~\ref{fig:architecture} and Table~\ref{tab:supported_skills}) and then input this embedding to the ANN search for retrieving relevant documents.

As the built-in Datastores, Wikipedia\footnote{The English
Wikipedia dump preprocessed by~\citet{karpukhin-etal-2020-dense}.} with the DPR encoder~\cite{karpukhin-etal-2020-dense}, PubMed\footnote{From the BioASQ8 edition~\cite{DBLP:conf/clef/NentidisKBKPVP20}.} and Bing web documents\footnote{From the MS MARCO dataset~\cite{bajaj2018ms}.} with the TAS-B encoder~\cite{DBLP:conf/sigir/HofstatterLYLH21} are supported. We plan to add more Datastores in the future.

\subsection{Explainability} \label{sec:explain_api}
Recently, many interpretability techniques to understand black-box neural models such as influence functions and input/token attribution methods~\cite{intersurvey21} have been introduced. Most of these techniques provide only local explanations and require access to the back-propagation function. One exception is \emph{CheckList} \cite{ribeiro2020accuracy}, which is a type of behavioural testing that treats models---in our case Skills---as black-boxes and compares their behaviour against the expected one. 
This is achieved by \textit{unit tests} designed by the end-users or the system experts. Two most common test types are \textbf{M}inimum \textbf{F}unctionality \textbf{T}est (MFT) and \textbf{INV}ariance (INV) as shown in Table~\ref{tab:expl_examples}. MFTs are designed to measure a capability (e.g., \textit{Taxonomy} capacity of matching object properties to categories) via specifying the expected behaviour (e.g., ``tiny'' in Table~\ref{tab:expl_examples}). INVs tests are similarly refined for capabilities (e.g., robustness under spelling errors in question), however the expected behaviour is already known, i.e., the answer should remain the same.  
We adapt the machine comprehension tests from \citet{ribeiro2020accuracy} for behavioural testing of our Skills. 
In our current setup, the tests for all the deployed Skills are curated manually, saved in as JSON file and made available via the UI. The test results are shown on demand via a separate tab (\S \ref{ssec:expl}).

\begin{table}
	\small
	\centering
	\resizebox{7.5cm}{!}{
    \begin{tabular}{p{\columnwidth}}
    	\toprule
    	\textbf{M}inimum \textbf{F}unctionality \textbf{T}est (MFT)-\textit{Taxonomy} \\
    	\midrule
        \textbf{C:} There is a \textcolor{red}{tiny} purple box in the room.\\
        \textbf{Q:} What size is the box?\\
        \textbf{Test:} Check if the prediction is \textcolor{red}{tiny}\\
        \midrule 
        \textbf{INV}ariance-\textit{Robustness}\\
        \midrule
        \textbf{C:} ...Newcomen designs had a duty of about 7 million, but most were closer to 5 million....\\
        \textbf{Q:} What was the ideal [\textcolor{red}{duty}->\textcolor{blue}{udty}] of a Newcomen engine? \\
        \textbf{Test:} Check whether the prediction changes or not.\\
        \bottomrule
    \end{tabular}}
    \caption{Examples for two most common test types. \textbf{Top:} Minimum Functionality Test~(MFT), \textbf{Bottom:} Invariance Test (INV). \textbf{C:} refers to context and \textbf{Q:} is the question.}
    \label{tab:expl_examples}
    \vspace{-5mm}
\end{table}
\vspace{-1mm}
\subsection{User Interface}\label{sec:user_interface}

We host \square as a web application built with VueJS\footnote{\url{https://vuejs.org}} to make it easily accessible to researchers. Once a Skill has been created by a user (\S \ref{sec:add_skill}) it can be added, edited, and deleted in the Skill management section of the application in the “My Skills” menu. For each Skill, its \textsc{URL}, metadata, requirements for context, and visibility can be adjusted (see Appendix Fig.~\ref{fig:skill-management}). The functionality of the user interface is split into QA and explainability.

\begin{table}
\centering
\resizebox{\linewidth}{!}{
\begin{tabular}{lp{0.42\linewidth}}
\toprule
 \textbf{Training Dataset for Models} & \textbf{Domain} \\ \midrule
\textbf{Text Generation (Abstractive QA)} &  \\
NarrativeQA \citep{kocisky2018narrativeqa} & Stories \\ \midrule
\textbf{Span Extraction (Extractive QA)} & \\
BioASQ \citep{tsatsaronis2015overview} & Biomedical \\
 DROP \citep{Dua2019DROP} & Wikipedia \\
DuoRC \citep{saha2018duorc} & Movies \\
 Natural Questions \citep{kwiatkowski2019natural} & Wikipedia \\
 NewsQA \citep{trischler2017newsqa} & News \\
 Quoref \citep{dasigi2019quoref} & Wikipedia \\
 SQuAD 1.1 \citep{rajpurkar2016squad} & Wikipedia \\
 SQuAD 2.0 \citep{rajpurkar2018know} & Wikipedia \\
 TriviaQA \citep{joshi2017triviaqa} & Wikipedia, Web \\ \midrule
\textbf{Text Classification (Multiple-Choice QA)} & \\ 
BioASQ \citep{tsatsaronis2015overview} & Biomedical \\
 BoolQ \citep{clark2019boolq} &  Wikipedia \\ 
 CommonsenseQA \citep{talmor2019commonsenseqa} & - \\
CosmosQA \citep{huang2019cosmos} & Personal Narratives \\
 MultiRC \citep{khashabi2018looking} & Fiction, Textbook, Wikipedia, News, etc. \\
 Quail \citep{rogers2020getting} & Fiction, News, Blogs, User Stories \\
 Quartz \citep{tafjord2019quartz} & Relationships \\
 RACE \citep{lai2017race} & News, Stories, Ads, Biography, Philosophy \\
 SocialIQA \citep{sap2019social} & Social Interactions \\ 
 \midrule
 \textbf{Query Embedder (Retrieval)} & \\ 
 Natural Questions~\citep{kwiatkowski2019natural} & Wikipedia \\
 MS MARCO~\citep{bajaj2018ms} & Bing Web Docs. \\
 \bottomrule
\end{tabular}
}
\caption{Available Models fine-tuned on various datasets upon the release of \square.}

\label{tab:supported_skills}
\vspace{-4mm}
\end{table}
\vspace{-2mm}
\paragraph{QA Interface.} The QA section of the user interface provides access to the Skill by allowing the user to enter their question and optionally a context. Public Skills are accessible to everyone while private Skills require the user to be signed in. The UI provides distinct visualizations depending on the selected Skill type. For extractive Skills, e.g., SQuAD~\citep{rajpurkar2016squad}, a document and multiple spans are returned and ranked by the model's confidence. In this setup, we also provide the option to show the span highlighted in its position in the document (see Fig.~\ref{fig:screenshot-question}). Categorical Skills, e.g., BoolQ~\citep{clark2019boolq}, show an interface with boolean output scores (see Appendix Fig.~\ref{fig:qa-boolq}). A multiple-choice Skill requires multiple options separated by newlines in the context field. These are then ranked and returned with their respective scores (see Appendix Fig.~\ref{fig:qa-multiple}). 
When multiple Skills are selected, the user can see and compare their outputs side-by-side and better understand their behavioural differences. 

\paragraph{Explainability Interface.} A Skill selector is provided at the top which allows users to visualize and compare the results of the CheckList machine reading tests for the selected Skills. A list of tests with their name, type, capability, and failure rate is shown. The list can be expanded for a detailed description along with a small number of failed examples with their questions, context, and predictions.

\section{Use Cases}

\subsection{Skill Publishing} \label{sec:add_skill}

A major contribution of our platform is to support developers creating their own Skills. This allows practitioners to easily make their research publicly available, without having to take care of engineering heavy topics such as infrastructure, web development and security. To publish a new Skill, developers need to implement a single function that defines the question answering pipeline. They are provided with utility functions that facilitate interacting with other components such as the Datastores, Models and the UI. A code snippet implementing a Skill is given in Appendix \ref{sec:predict_function}. 

Allowing developers to implement their own Skills enables us to greatly extend the system to have stronger models. For instance, multiple Datastores with potentially different retrieval methods can be combined to find complementary background knowledge, e.g., from Wikipedia and biomedical articles. Similarly, different models could be used to precisely answer a diverse set of questions that might require different capabilities, such as answerability \citep{rajpurkar2018know}, numerical \citep{Dua2019DROP} or multi-hop \citep{yang-etal-2018-hotpotqa} reasoning. Once a developer creates their Skill, it can be added to \square via the UI. The Skill developer can further make the Skill publicly available. \par
Allowing the community to implement Skills comes with a technical challenge such as deploying unreliable code on our servers. We therefore allow three different ways of hosting Skills. (1) First, Skills can be hosted directly on \square. For this, a pull request for the new Skill should be submitted to our public repository, which can then be added to the system upon a code review. While processing the submitted Skill requires a human in the loop, this option simplifies the hosting process for the Skill developer. (2) Second, in order to provide an option to make Skills instantly and independently available, we also allow Skills to be hosted on third party cloud platforms such as Amazon Web Services, Google Cloud and Microsoft Azure. All these cloud providers allow to easily host a lightweight function that can be used by \square. (3) Lastly, we allow developers to host Skills on their own hardware. The only requirement is that the Skill needs to be publicly accessible. In the latter two cases, developers will still have access to \square's components (e.g., Datastores and Models), but the Skill itself will run on the cloud or on other hardware.
For quick development of Skills we recommend using options (2) and (3). For long-term availability and usage of a Skill, adding it via the public github repository is recommended.  We provide extensive documentation for all possibilities to host Skills.\footnote{{\small \url{https://square.ukp-lab.de/docs/}}}\par

\begin{table*}[t]
	\small
	\centering
	\resizebox{1\textwidth}{!}{
\renewcommand{\arraystretch}{1}
    \begin{tabular}{lp{4.4cm}llll} 
        \toprule
        & \textbf{Supported Models} & \textbf{Retrieval} & \textbf{QA Types} & \textbf{Expl.} & \textbf{Ext.} \\
        \midrule
        Haystack & HF Transformers & sparse, dense & EX, AB &  $\times$ & $\times$ \\
        \citet{dibia2020neuralqa} & HF Transformers & sparse & EX  & \checkmark &  $\times$ \\
        \citet{karpukhin-etal-2020-dense} & DPR & dense & EX &  $\times$ & $\times$ \\
        \citet{khashabi2020unifiedqa} & T5 & $\times$ & EX, AB, MC, YN  & $\times$ & $\times$ \\
        \midrule
        \square & HF Transformers, ONNX, adapters & sparse, dense & EX, AB, MC, YN &  \checkmark & \checkmark \\
        \bottomrule
    \end{tabular}}
    \caption{
    Qualitative comparison of \textsc{UKP-SQuARE} to previous works. HF: HuggingFace, Expl.: Explainability component, Ext.: Extensible by the end-user
    , EX: Extractive, AB: Abstractive, MC: Multiple-choice, YN: Yes/No 
    }
    
    \label{table:prev_comp}
\end{table*}

\subsection{Skill Querying} 
Once a developer makes their Skill public in \square, other users can obtain answers from it. Upon release of the system, we make a wide range of question answering Skills available. These span over different QA formats (extractive, multiple-choice, abstractive), setups (open-domain, machine reading comprehension) and to different domains (wikipedia, web, biomedical, etc.). The list of available models for different formats is given in Table \ref{tab:supported_skills}. This allows the public to test current state-of-the-art question answering models. Moreover, researchers can use it for qualitative analysis, for example to discover potentials biases, strengths or weaknesses in models by behavioural testing. Furthermore, we support querying multiple Skills at the same time. This is particularly useful to compare capabilities of different models. For example see Fig.~\ref{fig:screenshot-question}, where two open domain, extractive Skills can be compared.

\subsection{Behavioural Testing of Skills}
\label{ssec:expl}

The users can choose the Skill they want to investigate from the drop-down menu. The selected Skill can be analyzed standalone or alongside two different compatible Skills. 

The tests are displayed showing the Skill failure rate and the failed examples can be viewed by clicking on the `Expand` button. An examplary visualization for negation and coreference testing of SQuAD Skills is given in Appendix Fig.~\ref{fig:checklist_ui}. For replacement tests, e.g., where names are perturbed, colored markers are used to highlight how the input was modified for the test. This allows the user to quickly identify changes the Skill could not handle. To analyze or process a Skill’s test performance in more detail, a full JSON report of all test examples can be downloaded. 

\section{User Study}
\label{sec:user_study}
We evaluate the usability of our system by conducting a pilot attitudinal user study with five participants. We recruited graduate students, our main target user group, and instructed them to compare and analyze several Skills. We provided them with a list of predefined questions to input into the system to help them use it. After the students used the system we asked them several questions to discover whether they understood every element of the interface effortlessly (i.e., the input and the output of the Skills, the list of behavioral cards of the Skills, and their specific contents). All users understood the input and output of the Skills and stated that the interface allows them to compare the Skills effortlessly. They also stated that the behavioral cards of the explainability component are useful to analyze the strong and weak points of the models and could help develop new Skills. However, most of them could not understand them in a glimpse. Hence, we will improve the presentation of these cards in a future update. Appendix \ref{app:user_study} provides the list of questions and responses. To finish the study, we employed the System Usability Scale (SUS) questionnaire \citep{brooke1996sus} to quantitatively assess the global usability of the system. The average score is 70 out of 100, which refers to a ``good usability" \citep{UIUX-Trend}.

\section{Related Work}

A qualitative comparison with similar frameworks is given in Table~\ref{table:prev_comp}. 
The closest work to ours is Haystack, which is an open-source and scalable framework for building search systems over large document collections. Although it supports both sparse and dense retrieval techniques, models from the Huggingface (HF), and different QA types (abstractive and extractive) it lacks support for faster \texttt{ONNX} or memory efficient \texttt{adapter} models. Furthermore, it has to be set up by the users on their own infrastructure which requires technical expertise and sufficient hardware resources. \citet{dibia2020neuralqa} introduce NeuralQA, an interactive tool for QA that leverages the benefits of sparse retrieval along with the HF reader models. However, NeuralQA is limited to extractive QA. \citet{karpukhin-etal-2020-dense} provide a simple user interface that employs efficient dense retrieval but only support models for open-domain QA. Finally, UnifiedQA~\cite{khashabi2020unifiedqa} provides a demo page\footnote{\url{https://unifiedqa.apps.allenai.org/}} that employs a custom T5 based model trained on a wide range of QA datasets, hence supports a variety of QA formats. However, (1) it lacks the retrieval component, (2) is not scalable (to include different model formats), and (3) is not flexible (not possible to use models with different retrieval techniques). Unlike other previous systems, \square is dynamically extendable allowing users to easily contribute with new Skills. Finally, except from gradient-based explanations in \citet{dibia2020neuralqa}, none of the systems have an explainability component.

\section{Conclusion and Future Work}
We introduce the \square platform that enables researchers and developers to study and compare QA pipelines, i.e., Skills, that comprises a selection of Datastores, retrieval mechanisms and reader models. The platform enables querying existing public Skills, as well as implementing custom ones using \square's microservices and utility functions that support a large collection of model types and Datastores. Furthermore, users can simultaneously query multiple Skills, and analyze them through integrated behavioural tests.

Our architecture is scalable and flexible to incorporate most of the latest developments in the QA domain. Future versions will include automated deployment of custom models and Datastores, automated Skill selection by incorporating previous works~\cite{metaqa21,tweac21} and increasing the number of supported Datastores (e.g., wikidata, \citeauthor{10.1145/2629489}, \citeyear{10.1145/2629489}). We also plan to incorporate specialized models (e.g., using graph encoders, \citeauthor{ribeiro-etal-2021-structural}, \citeyear{ribeiro-etal-2021-structural}), structured reasoning approaches (\citeauthor{yasunaga-etal-2021-qa}, \citeyear{yasunaga-etal-2021-qa}) and interpretability techniques such as saliency maps~\cite{LiCHJ16}.

\label{sec:bibtex}

\section*{Ethics and Broader Impact Statement}

\paragraph{Data}

This work does not generate new data. All datasets employed in used to construct Skills as described in \S\ref{sec:model_api}, \S\ref{sec:datastore_api}, and Table \ref{tab:supported_skills}. The datasets are well-known to be safe for research purposes and do not contain any personal information or offensive content. We comply with the licenses and intended uses of each dataset. The licenses of each dataset can be seen in Appendix \ref{sec:dataset_lic}.
\paragraph{Intended Use.}
The intended use of \square is i) bringing different QA components together to share them as a skill with the rest of the world and ii) the analysis of these Skills. Our platform allows NLP practitioners to share their Skills with the community removing technical barriers such as configuration and infrastructure so that any person can reuse these models. In addition, users can analyze the available Skills through behavioral tests and compare them thanks to a user-friendly UI. This has a straightforward benefit for the research community (i.e., reproducible research and analysis of prior works), but also to the general public because \square allows them to run state-of-the-art models without requiring them any special hardware and hiding complex settings such as virtual environments and package management.

\paragraph{Potential Misuse.}
Our platform makes use of Skills uploaded by the community. However, this current version does not incorporate any mechanism to ensure that these models are fair and without bias. Nonetheless, \square includes a module for explainability that uses CheckLists \citep{ribeiro2020accuracy} to analyze the strong and weak points of the Skills and to detect their biases and unfair content. Thus, we currently delegate the fairness checks to the authors of the models. We are not held responsible for errors, false, or offensive content generated by the Skills. Users should use them at their discretion.

\paragraph{Environmental Impact.}
Since \square empowers the community to run publicly available Skills on the cloud, it has the potential to reduce CO\textsubscript{2} emissions from retraining previous models to make the comparisons needed when developing new models.

\paragraph{User Study.} The participants are junior graduate students recruited on a voluntary basis. They are not part of this work, and never saw the user interface before the study. Before starting the study, they were given detailed instructions on the goals and scope of the study, and how the data was going to be used. Only non-personal data was recorded.

\section*{Acknowledgements}
We thank Jan-Christoph Klie for his insightful feedback and suggestions on a draft of the paper and the project. We thank Richard Eckart de Castilho for advice on the general infrastructure and Nandan Thakur and Hossain Shaikh Saadi for their preliminary work on the project. We also thank Serkan Bayraktaroğlu for designing our logo. \par 
This work has been funded by: (i) the German Research Foundation (DFG) as part of the UKP-SQuARE project (grant GU 798/29-1), (ii) the DFG as part of the QASciInf project (GU 798/18-3), (iii) the DFG within the project “Open Argument Mining” (GU 798/25-1), associated with the Priority Program “Robust Argumentation Machines (RATIO)” (SPP-1999), (iv) the DFG-funded research training group ``Adaptive Preparation of Information form Heterogeneous Sources'' (AIPHES, GRK 1994/1), (v) the European Regional Development Fund (ERDF) and the Hessian State Chancellery – Hessian Minister of Digital Strategy and Development under the promotional reference 20005482 (TexPrax), and (vi) the LOEWE initiative (Hesse, Germany) within the emergenCITY center.

\bibliography{main}
\bibliographystyle{acl_natbib}

\appendix
\clearpage

\section{Skill Implementation}\label{sec:predict_function}
The code below implements an open-domain, extractive QA Skill. First, a set of utility classes are loaded and initialized for facilitating interaction with \square's Models and Datastore components (lines 1-5). Next, in the \texttt{predict} function, the Datastores are queried for retrieval. The Datastores component takes the user query, the datastore (Wikipeida snapshot from Natural Questions) and what index to use (dense, based on DPR) as input and returns the top documents. From these results, the document text and respective scores are extracted (lines 11-17).
Subsequently, the query and the top documents are passed to an to the Models component for span extraction. In this implementation, a BERT base model with a adapter trained on SQuAD V2.0 is used (lines 21-30). Finally, the top answers are returned (lines 32-36).

\begin{listing*}[!b]
\begin{minted}
[
% frame=lines,
framesep=1mm,
baselinestretch=1,
% bgcolor=LightGray,
fontsize=\small,
linenos,
xleftmargin=1.5em,
]
{python}
from square_skill_api.models import QueryOutput, QueryRequest
from square_skill_helpers import ModelAPI, DataAPI

model_api = ModelAPI()
data_api = DataAPI()

async def predict(request: QueryRequest) -> QueryOutput:
    
    # Dense document retrieval using the Datastores 
    # on a Wikipedia snapshot with DPR embeddings
    data_api_output = await data_api(
        datastore="nq",
        index_name="dpr",
        query=request.query,
    )
    context = [d["document"]["text"] for d in data_api_output]
    context_score = [d["score"] for d in data_api_output]
    
    # Answer extraction from the top document using the Model API
    # using bert-base-uncased base model with SQuAD2.0 adapter
    model_api_request = {
      "input": [[request.query, c] for c in context],
      "task_kwargs": {"topk": 1},
      "adapter_name": "qa/squad2@ukp",
    }
    model_api_output = await model_api(
        model_name="bert-base-uncased",
        pipeline="question-answering",
        model_request=model_api_request,
    )

    return QueryOutput.from_question_answering(
        model_api_output=model_api_output,
        context=context,
        context_score=context_score
    )
\end{minted}
\caption{Example Implementation of an open-domain, span extraction Skill.}
\label{predict-fn}
\end{listing*}
\clearpage

\section{Dataset Licences}
\label{sec:dataset_lic}
Table \ref{table:license} shows the license of each dataset. In the case of RACE, the authors did not provide any license but specified that it can only be used for non-commercial research purposes. In the case of the other datasets without any specified license, the authors did not provide any license, but the datasets are freely available to download and use in a research context. BioASQ is available by Courtesy of the U.S. National Library of Medicine.

\begin{table}[h]
\begin{center}
\begin{tabular}{lp{0.5\linewidth}}
\toprule
\textbf{Dataset}       & \textbf{License}      \\
\midrule
NarrativeQA   & Apache 2.0   \\
BioASQ   & National Library of Medicine Terms and Conditions   \\
DROP          & CC BY-SA 4.0 \\
DuoRC         & MIT          \\
Natural Questions   & MIT          \\
NewsQA         & MIT          \\
Quoref         & CC BY 4.0     \\
SQuAD 1.1      & CC BY-SA 4.0         \\
SQuAD 2.0    & CC BY-SA 4.0          \\
TriviaQA     & Apache 2.0         \\
BoolQ         & CC BY-SA 3.0 \\
CommonSenseQA & NA           \\
CosmosQA         & NA          \\
MultiRC         & NA          \\
Quail         & NA          \\
Quartz         & NA         \\
RACE          & NA           \\
SocialIQA          & NA           \\
MS MARCO         & CC BY 4.0          
\\ \bottomrule
\end{tabular}
\caption{License of each dataset.}
\label{table:license}
\end{center}
\end{table}

\section{Questions of the User Study}
\label{app:user_study}
Table \ref{tab:questions_user_study} contains the answers of the participants of the user study (\S \ref{sec:user_study}) to each question we asked to evaluate their understanding of the interface. 

\newpage 
\begin{table}[!htp]
\begin{tabular}{p{0.66\linewidth}c}
\\\toprule
\textbf{Question}                               & \textbf{Avg. Ans.} 
\\ \midrule
SQuARE provides a user interface that allows me to tell the difference between both Skills &  4.4\\
I understand in a glimpse each card.            &   2.6 \\
I can get a quick overall view of the weak points of the skill. & 3.8  \\
The examples of each CheckList item are useful. &  4.4                
\\\bottomrule
\end{tabular}
\caption{List of questions to understand the usefulness of the system. 1 represents "strongly disagree" and 5 represents "strongly agree."}
\label{tab:questions_user_study}
\end{table}

\section{User Interface}
\label{app:qa}

\begin{figure*}[h]
    \centering
    \includegraphics[width=0.9\textwidth]{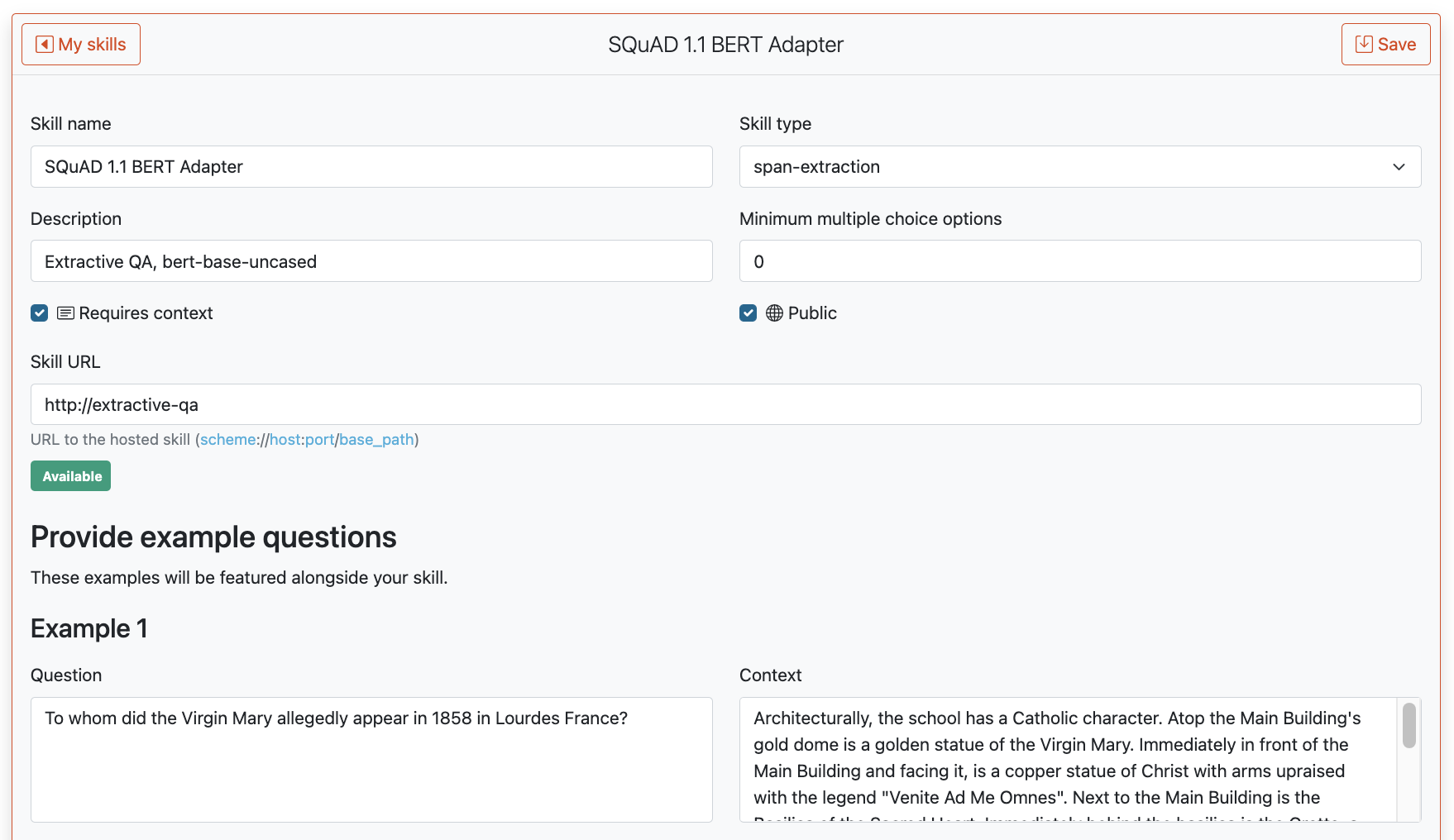}
    \caption{User interface for managing a Skill.}
    \label{fig:skill-management}
\end{figure*}

\begin{figure*}[!htp]
    \centering
    \includegraphics[width=0.9\textwidth]{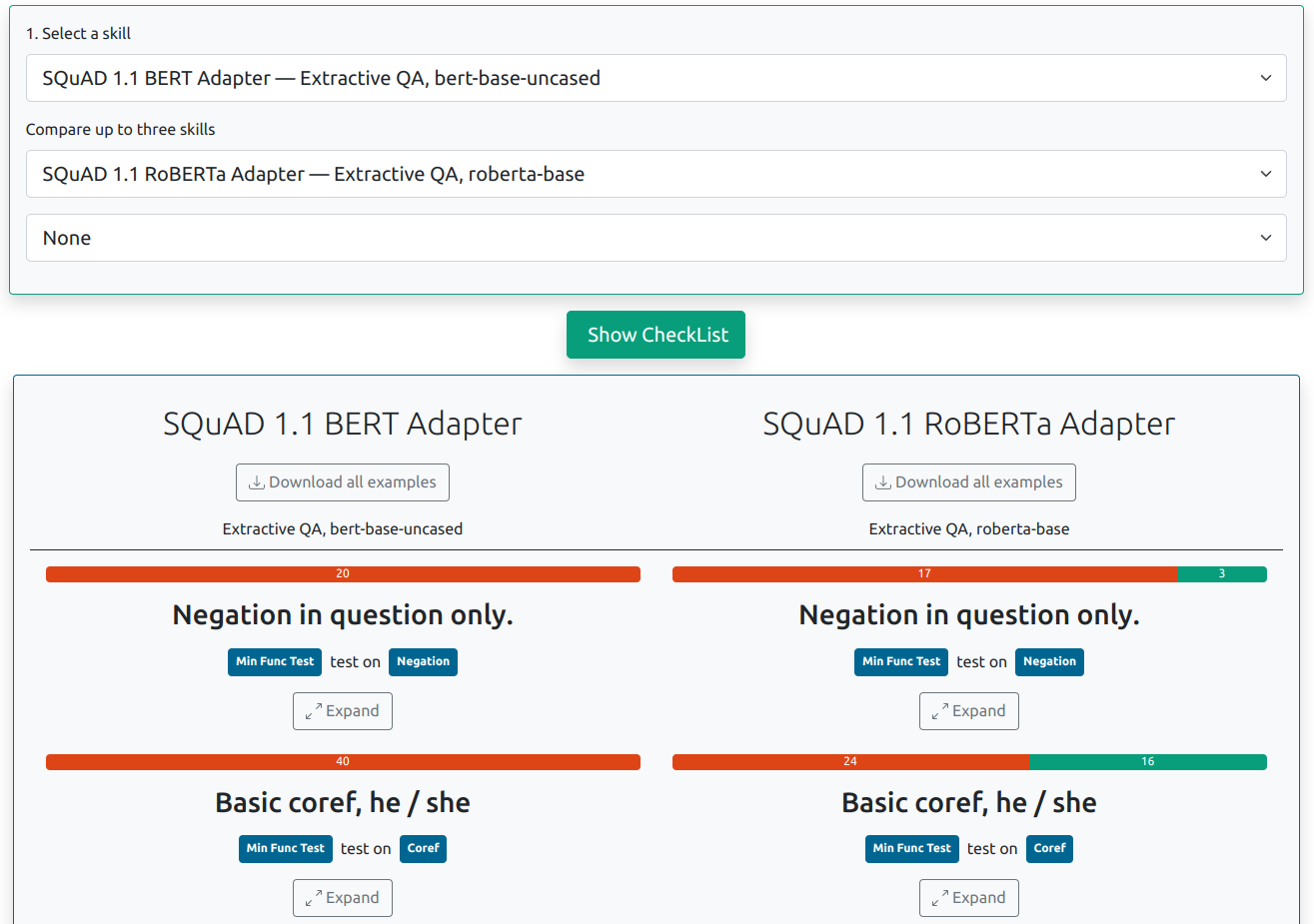}
    \caption{User interface for behavioural tests from CheckList.}
    \label{fig:checklist_ui}
\end{figure*}

\begin{figure*}[!htp]
    \centering
    \includegraphics[width=0.9\textwidth]{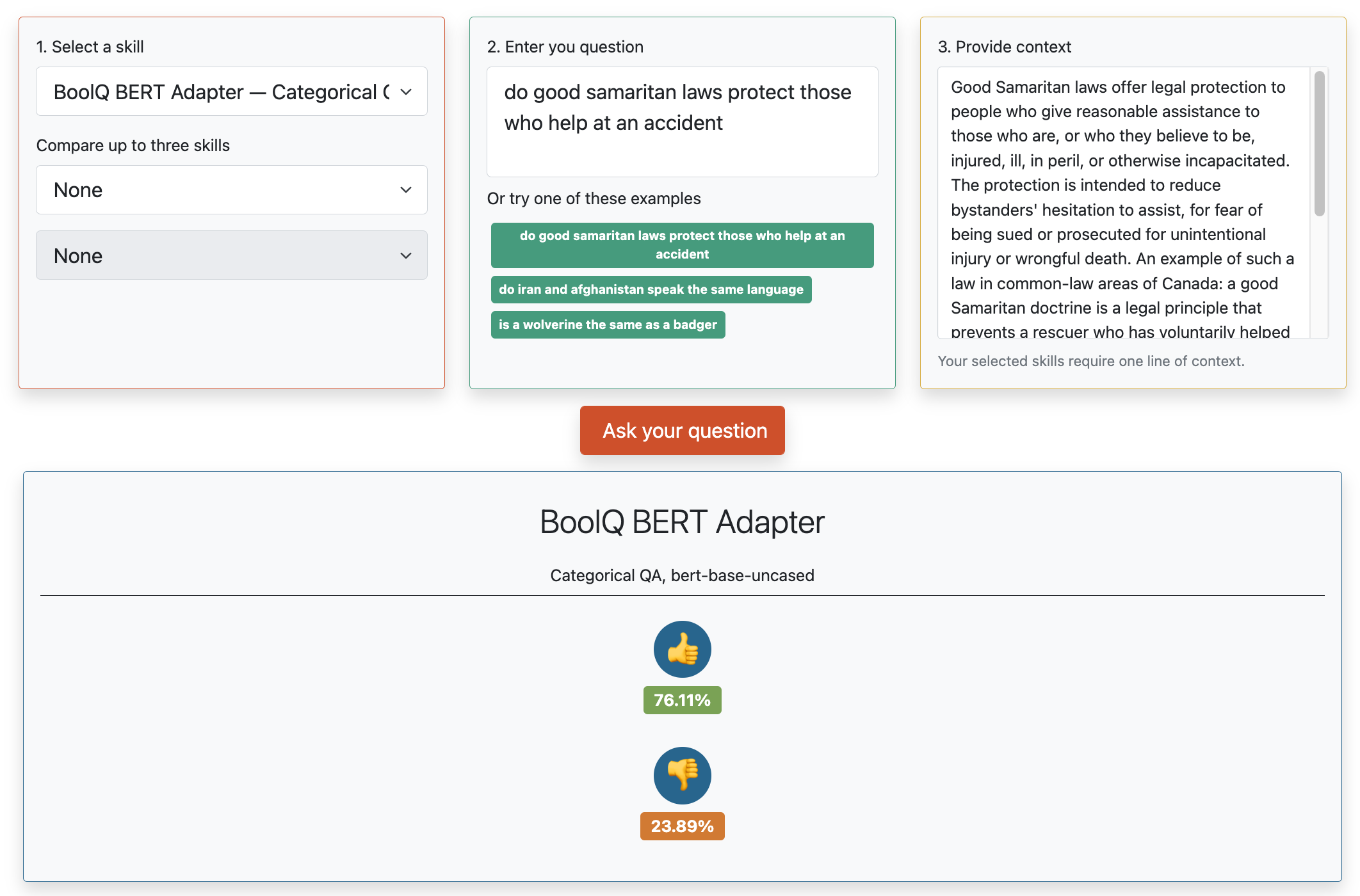}
    \caption{User interface for visualizing categorical Skill results.}
    \label{fig:qa-boolq}
    \vspace{16.2mm}
\end{figure*}

\begin{figure*}[!h]
    \centering
    \vspace{15mm}
    \includegraphics[width=0.9\textwidth]{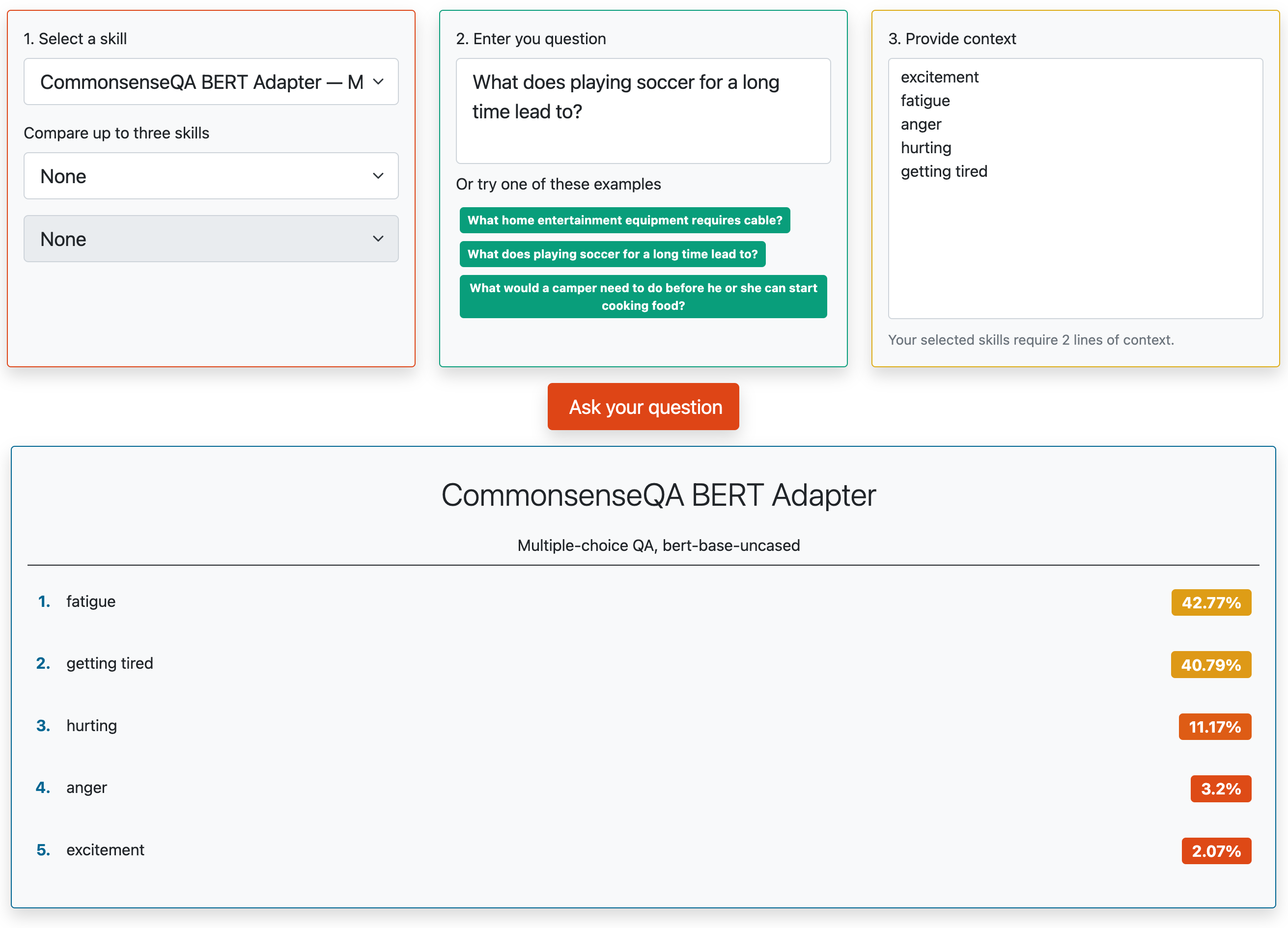}
    \caption{User interface for visualizing multiple choice Skill results.}
    \label{fig:qa-multiple}
    \vspace{-4mm}
\end{figure*}

UI screenshots for visualizing categorical and multiple choice Skill results are given in Fig.~\ref{fig:qa-boolq} and~\ref{fig:qa-multiple} respectively. In Fig.~\ref{fig:skill-management} the UI for managing a Skill is shown. Navigating through behavioural test results is given in Fig.~\ref{fig:checklist_ui}.

\clearpage

\end{document}